\documentclass[conference]{IEEEtran}
\IEEEoverridecommandlockouts
\usepackage{cite}
\usepackage{amsmath,amssymb,amsfonts}
\usepackage{algorithmic}
\usepackage{graphicx}
\usepackage{textcomp}
\usepackage{xcolor}
\usepackage{float}
\usepackage{graphicx}
\usepackage{booktabs,ragged2e} 
\def\BibTeX{{\rm B\kern-.05em{\sc i\kern-.025em b}\kern-.08em
    T\kern-.1667em\lower.7ex\hbox{E}\kern-.125emX}}
\begin{document}

\title{Simulations and Advancements in MRI-Guided Power-Driven Ferric Tools for Wireless Therapeutic Interventions
}

\author{\IEEEauthorblockN{\textsuperscript{} Wenhui Chu $^{\dag}$,  Aobo Jin $^{\dag}$,  Hardik A. Gohel $^{\dag}$}\\

\IEEEauthorblockA{$^{\dag}$\textit{Dept. of Computer Science, University of Houston-Victoria,  Victoria, USA.} \\ChuW1@uhv.edu,  Jina@uhv.edu, gohelh@uhv.edu}

}

%\author{\IEEEauthorblockN{\textsuperscript{} Wenhui Chu $^{\dag}$, Giovanni Molina $^{\dag}$, Nikhil V. Navkar $^{\dag}^{\dag}$, Christoph F. Eick $^{\S}$, }{Aaron T. Becker $^{\P}$, Panagiotis Tsiamyrtzis $^{\ast}$, Nikolaos V. Tsekos $^{\dag}$}\\

%\IEEEauthorblockA{$^{\dag}$\textit{MRI Lab, Dept. of Computer Science, University of Houston, Houston, USA.} \\wchu@uh.edu, gemolinaramos@uh.edu, nvtsekos@central.uh.edu}
%\IEEEauthorblockA{$^{\dag}^{\dag}$\textit{Dept. of Surgery, Hamad Medical Corporation, Doha, Qatar.} \\nnavkar@hamad.qa}
%\IEEEauthorblockA{$^{\S}$\textit{ DAIS Lab, Dept. of Computer Science, University of Houston, Houston, USA.} \\ceick@uh.edu

%\IEEEauthorblockA{$^{\P}$\textit{Dept. of Electrical and Computer
%Engineering, University of Houston, Houston, USA. } \\atbecker@uh.edu}
%\IEEEauthorblockA{$^{\ast}$\textit{Dept. of Statistics, Athens University of Economics and Business, Greece.} \\
%pt@aueb.gr}

%}

\maketitle

\begin{abstract}

Designing a robotic system that functions effectively within the specific environment of a Magnetic Resonance Imaging (MRI) scanner requires solving numerous technical issues, such as maintaining the robot's precision and stability under strong magnetic fields. This research focuses on enhancing MRI's role in medical imaging, especially in its application to guide intravascular interventions using robot-assisted devices. A newly developed computational system is introduced, designed for seamless integration with the MRI scanner, including a computational unit and user interface. This system processes MR images to delineate the vascular network, establishing virtual paths and boundaries within vessels to prevent procedural damage. Key findings reveal the system's capability to create tailored magnetic field gradient patterns for device control, considering the vessel's geometry and safety norms, and adapting to different blood flow characteristics for finer navigation. Additionally, the system's modeling aspect assesses the safety and feasibility of navigating pre-set vascular paths. Conclusively, this system, based on the Qt framework and C/C++, with specialized software modules, represents a major step forward in merging imaging technology with robotic aid, significantly enhancing precision and safety in intravascular procedures.

\end{abstract}

\begin{IEEEkeywords}
MRI, medical imaging, robot-assisted devices, magnetic field gradient
\end{IEEEkeywords}

\section{Overview}

The use of robot-assisted systems in minimally invasive surgery (MIS) is growing, leading to marked improvements in the precision and accuracy of these procedures. Further development of robotic technology within MIS is expected to enhance the capabilities of these systems and provide solutions to the complexities involved in advanced surgical techniques. The focus on developing MRbots aligns with this trend, as these devices are designed to navigate the human body with minimal invasiveness.

The evolution of micro- and nanorobotics has led to the creation of navigable or actuated magnetic spheres, which can be precisely controlled within the body. For practical applications, especially in the biomedical field, it is essential for operators to have real-time imaging and tracking capabilities of miniature robots while they perform designated tasks within the human body [1]. By combining magnetic actuation with advanced imaging modalities, it becomes possible to track the MRbots within the body. This capability is crucial for implementing closed-loop servo control systems, where the MRbots can follow pre-planned trajectories with high accuracy.

The small size of capillaries, which permits only individual blood cells to pass, highlights the benefit of the compact design of smaller MRbots. Their reduced size allows them to traverse through tighter spaces within the body, including small blood vessels like capillaries, thereby broadening their operational scope for more focused and precise medical interventions. Existing data suggests that reducing variability and improving perioperative care protocols could enhance surgical outcomes and lower costs for patients undergoing operations [2]. Ongoing studies in this field are investigating various ways to use MRbots in minimally invasive surgery (MIS), aiming to improve surgical efficacy, reduce patient discomfort, and shorten recovery times. This growing area in robotics research emphasizes the immense potential in combining robotics, imaging technology, and advanced surgical techniques to advance medical care.

 Medical microrobots driven by magnetic fields can be operated using current clinical MRI systems [3]. This approach serves dual functions: it allows for the imaging and monitoring of the microrobots and also controls their movement and propulsion [4][5].  This method addresses the challenge of conducting surgery in areas with continuous motion, such as the heart.

Magnetic Resonance Navigation (MRN) presents the possibility of dynamically directing drug particles and cells to specific locations within the body [5].  In MRN, drug-loaded microcarriers, embedded within a magnetic sphere, are directed through the vascular network from an injection point to a specific target region. This precise targeting is particularly beneficial for delivering therapeutics to tumors, as it increases the efficacy of the treatment while simultaneously reducing secondary toxicity effects. This approach represents a significant advancement in the field of targeted therapy, offering a more efficient and less invasive method for treating various medical conditions [6].

Our research
is centered on advancing the field of medical imaging and robotics through the development of an innovative computational system that enhances the effectiveness and safety of MRI-driven, robot-assisted devices in vascular interventions. A significant feature is its ability to process MR images for mapping vascular networks, creating virtual pathways to prevent vessel damage during procedures. It leverages vessel geometric features and adheres to MRI safety protocols to craft precise magnetic field gradient patterns for applicator control, while also accommodating varying blood flow characteristics for individualized guidance. The paper is organized as follows: it begins with the approach and procedures section, which discusses the rationale behind selecting dynamic flow and analyzes the Proportional Integral Derivative (PID) controller strategy being examined. This is followed by section III, where the conceptual framework is elaborated. Section IV presents the modeling and experimental results. Finally, the paper concludes with a section focusing on discussions and final remarks.

\section{APPROACH AND PROCEDURES}

Several computational analyses have been conducted on the flow dynamics around 3D microbots within channel structures that simulate the vascular system. These studies aim to comprehend how the flow field changes when a microbot is introduced into fluid-filled channels [7]. Blood vessels serve as conduits for distributing blood to various tissues in the body, forming two separate circulatory loops that originate and conclude at the heart. The pulmonary circuit moves blood from the right ventricle to the lungs and back to the left atrium, while the systemic circuit circulates blood from the left ventricle to tissues throughout the body before returning it to the right atrium. Blood vessels are categorized into three types based on their structure and role: arteries, capillaries, and veins [8]. The implication is clear: the smaller the MRbot, the more extensive the portion of the human body it can access. This is especially relevant for navigating the intricate network of capillaries.  

In practical terms, deploying such MRbots in the human circulatory system typically involves an untethered approach, meaning these devices operate wirelessly. This aspect presents a considerable technical challenge with current technologies, given the complexities of navigating the dynamic and varied environment of the cardiovascular system [9].
For efficient movement, these MRbots must be integrated seamlessly into the normal blood flow, which requires careful consideration of their size and propulsion mechanisms. Smaller MRbots have the advantage of a broader operational range, as they can navigate through smaller blood vessels, including capillaries, which are inaccessible to larger devices. The design and propulsion of these MRbots are thus critical factors in their effectiveness and the potential applications in medical procedures and treatments. To ensure precise movement, especially in the more complex, tortuous parts of the vessel network, we employ a Proportional Integral Derivative (PID) controller. This controller is crucial for adjusting and stabilizing the robot's trajectory.

Essential aspects that determine the maneuvering of MRbots are the nature of blood flow and the variable sizes of the targeted blood vessels, which greatly influence the drag force applied to the MRbots, hence setting the parameters for their functioning. Furthermore, the quantity of ferromagnetic particles embedded within the MRbots is tailored to suit particular medical uses, ensuring that every MRbot is meticulously crafted for its specific purpose in therapeutic or diagnostic procedures.

\section{CONCEPTUAL FRAMEWORK}

This study concentrates on a uniformly shaped spherical MRbot, tailored for accurate maneuvering inside the human body. The consistent production of gradient forces along the scanner's perimeter facilitates force generation at any bodily location, providing precise temporal control over the MRbot's movements. This characteristic is particularly beneficial for navigating through narrow blood vessels, where the impact of the MRbot's buoyancy and weight is minimal in comparison to the drag force, which is substantially more pronounced. Another important factor is the pulsating pattern of blood flow within the arterial system. Close to the heart, where blood is ejected, the flow rate is significantly higher. However, as the blood travels further from the heart, its speed gradually diminishes, eventually slowing to rates measurable in centimeters and then millimeters per second. This disparity in blood flow speeds throughout the vascular system poses distinct challenges and possibilities for directing and managing the MRbot [10].

\subsection{Data Estimation Technique}

In our study, we highlight the importance of interpolation in the field of medical imaging, especially in the context of volumetric imaging. To tackle the issue of uneven data sampling, we utilize a specialized interpolation method called Piecewise Cubic Hermite Polynomials (PCHIP). Our method involves selecting a specific interval [a, b] and a function d that operates within this interval, transforming from [a, b] to the real numbers $\to$ $\Re$, together with its derivative d’. We subsequently employ a cubic Hermite spline, represented as s, to closely mimic the behavior of the function d across the interval [a, b]. Tortuosity is characterized as the most direct path through a medium, necessitating the calculation of the shortest geodesic paths. It's vital to acknowledge that tortuosity varies with direction. We depict it as a three-dimensional, parametric space curve, defined in Cartesian coordinates by \textit{K} = (x(t), y(t), z(t)), where each segment of the curve aligns with one of the Cartesian axes (X, Y, and Z). This formulation enables an accurate description of tortuosity in three-dimensional environments, essential for comprehending and maneuvering through intricate routes within any medium.

\begin{equation}
 \label{eq:curvature}
K =\frac{\sqrt{(z''y'-y''z')^2+(x''z'-z''x')^2+(y''x'-x''y')^2}}{{(x'^2+y'^2+z'^2)}^{\frac{3}{2}}}\qquad
\end{equation}

In our mathematical model, $x'$, $y'$, and $z'$ symbolize the first derivatives of the preferred path relative to a specific parameter, often time or the distance along the path. These derivatives represent the velocity components' rate of change along the X, Y, and Z axes. In a similar vein, $x''$, $y''$, and $z''$ are the second derivatives of the path, indicating the acceleration components along each axis. Analyzing both first and second derivatives is essential for comprehending the dynamics of motion along the path. This includes understanding factors like curvature and velocity change rate, which are crucial for accurate navigation and control in scenarios such as robotic path planning.

\subsection{Circulatory Speed} 

In our study, we recognize the heart's rhythmic beating as the origin of arterial blood pulsation, which leads to varying pressures on the vessel walls over time. To replicate this phenomenon, we employ a multi-faceted approach, each element offering its unique advantages and limitations [11][12].

Initially, we use zero-dimensional (0D) models to get a general, non-spatial perspective of blood flow dynamics. Next, we progress to one-dimensional (1D) models, concentrating on the longitudinal variations in blood pressure and flow [11]. These models are instrumental in examining wave movements and pressure differences within the vascular system.

For a deeper analysis, we turn to three-dimensional (3D) fluid-structure interaction methods. These sophisticated techniques accurately depict intricate flow patterns, wave transmissions, and the interplay between blood movement and vessel walls, crucial for a realistic understanding of blood flow dynamics.

Concerning drag force calculations for objects like MRbots moving through a fluid such as blood, we apply specific principles of fluid dynamics to determine the drag force impacting the sphere. This aspect is vital for the design and operation of micro-robots that must efficiently navigate the vascular system, considering the diverse forces encountered in different regions of the circulatory system.

\begin{equation}
 \label{eq:drag}
F_{drag}=\frac{1}{2} \cdot C_d\cdot p\cdot Re \cdot 
\left\vert V_{blood} - V_s\right\vert  
\end{equation}

The drag force exerted on an object moving through a fluid, such as blood, is calculated using key parameters including the drag coefficient ($C_d$), the density of the blood (\textit{p}), and the reference area ($R_e$). The drag coefficient ($C_d$) is a dimensionless number that quantifies the drag or resistance of an object in a fluid environment. It is determined by the shape of the object and the nature of the fluid flow around it. The blood density (\textit{p}) is considered as $1.025Kg/{m}^3$. This value is crucial for calculating the drag force, as the density of the fluid directly impacts the amount of resistance an object encounters while moving through it. The reference area ($R_e$) is another important factor in these calculations. It typically refers to the area of the object facing the flow of the fluid. This area is used as a reference point for calculating the forces exerted by the fluid on the object.

\subsubsection{Varying Blood Flow Patterns}

Our study is centered on examining blood flow under three varied pulsating scenarios: steady, normal, and accelerated heart rate flows [13]. Understanding how different heart rates influence the maneuverability and management of micro-robots in the vascular system is crucial in this research.

Magnetic field gradients, vital for directing the MRbot, are refreshed every 100 milliseconds. However, the MRbot's precise location is tracked more frequently, at intervals of Tp milliseconds. This interval, Tp, is adjusted based on the specific blood flow condition under examination. By tailoring Tp to steady, normal, and high heart rate scenarios, we can precisely observe the sphere's motion dynamics in each context.

For a comprehensive assessment, we conduct these experiments with distinct Tp intervals for each type of flow to thoroughly evaluate the sphere's performance under various physiological states. The findings from these tests are then analyzed and presented using the Qt platform, a prevalent tool in application development. This method enables us to effectively illustrate the MRbot's response to diverse blood flow conditions, offering significant insights for its usage in medical contexts.

\subsection{Speed of the Sphere}

Through the fusion of the trajectory and the intended velocity profile, we are able to proficiently steer the MRbot along diverse routes within the body. This involves modifying its speed and course in response to the fluctuating conditions of the vascular system. This strategy is vital for the effective deployment of MRbots in medical treatments and interventions. At each point of the trajectory, we determine the associated velocity vector, making sure that the MRbot maintains the appropriate speed and direction at all times. Such thorough computation of the velocity vector for each point is key for exact control of the MRbot's navigation, crucial in maneuvering through the intricate and ever-changing landscape of the human vascular system. We concentrate on the MRbot's trajectory, designed to reach a specified velocity, \textit{V(r)}, along its path \textit{P(r)} [14][15]. This trajectory transcends a mere route. It is a precisely devised vector function determining the velocity vector at each juncture along the path.

\begin{equation}
 \label{eq:velocity}
V(r) = \frac{V_0}{1+K/K_0}+\frac{R_s-R_{GC}}{{R_{0}}}\\
\end{equation}

 We use three constants: $V_0$, $K_0$, and $R_0$, to adjust the velocity profile of the MRbot. $V_0$ represents a baseline velocity, which serves as a reference point or starting velocity for the MRbot's movement. $K_0$ is a scaling factor that adjusts the velocity profile, allowing for changes in speed according to different conditions within the vascular network. $R_0$ is used to normalize or scale the radius of the MRbot in relation to the velocity profile. Additionally, The construction and movement specifications of the MRbot are meticulously determined. A primary factor is the MRbot's radius, labeled as $R_s$. This element is vital as it impacts the MRbot's capacity to traverse blood vessels of various sizes and also influences the drag force encountered. 

These constants play a crucial role in refining the MRbot's performance, guaranteeing its ability to adjust its velocity and navigational tactics suitably to the varied and ever-changing conditions of the human vascular system. Such control is critical for the accurate and secure functioning of MRbots in medical contexts, especially in situations that demand maneuvering through intricate vascular networks.

\subsection{Summary of MRI Gradient Features}

In our strategy for the safe navigation of MRbots in the vascular system, we introduce `Virtual Fixtures' (VF) – software-implemented guides that assist in steering. The application of VF is fundamental in two areas: ongoing tracking with path imaging and secure maneuvering. VF act as electronic navigational aids, steering the MRbot on its set course within the blood vessels. These aids ensure that the MRbot precisely reaches its target without harming the vessel walls. This technique is especially pertinent in our research on MRI-Guided and Powered Ferric Applicators for Tetherless Therapeutic Delivery. The trajectory is meticulously planned to adhere to a pre-set route, prioritizing both the safety of the procedure and the exact control of the MRbots. This plan accounts for the magnetic gradients produced by the MRI system, essential for directing and driving the MRbots. Additionally, the size of the ferromagnetic component inside the MRbot is chosen to correspond with the sizes of the intended blood vessels. This aspect is crucial to guarantee smooth passage of the MRbot through the vascular system to its end destination. Our method thus merges precise trajectory planning with the practical use of VF for the secure and efficient guidance of MRbots in medical procedures. In guiding MRbots, it's essential to accurately manage their path, positioning, and the gradients they encounter. Our approach necessitates creating three distinct setpoints for precise MRbot direction.

The initial setpoint is the establishment of coordinates outlining the path's central line, labeled as $P_t$. These coordinates act as crucial navigation points, directing the MRbot on its intended route within the vascular system.

The second setpoint pertains to the MRbot's speed. Here, the velocity at each point ($P_t$) is used as the benchmark. This ensures that the MRbot sustains the appropriate pace while navigating, adjusting to vascular system variations as necessary.

The third aspect of our control mechanism is the trajectory controller, comprising a PID regulator and a feedforward element. The PID regulator constantly fine-tunes the MRbot's course to minimize deviations and adhere to the set trajectory. Conversely, the feedforward component generates optimal control inputs based on the established trajectory and velocity setpoints.

Velocity discrepancies, critical for precise control, are determined by comparing the actual and target velocities at each path point. Continuously correcting these deviations enables the MRbot to traverse the human vascular system's intricate and dynamic milieu with great accuracy, thereby improving the safety and efficiency of medical procedures.

\begin{equation}
 \label{eq:error}
Error_v = V_c-V_s + K_r * (P_c-P_s)
\end{equation}

In our MRbot navigation control system, we focus on essential factors such as the current velocity ($V_c$), the velocity setpoint ($V_s$), and a crucial correction factor $K_r$, set at 0.7. This factor is pivotal in balancing position correction with velocity error correction within our algorithm.

The parameter $K_r$ = 0.7 determines how much the MRbot's positional errors are rectified relative to its velocity errors. This balance is vital for ensuring that the MRbot accurately follows its intended path (position precision) while maintaining the correct speed (velocity precision).

\begin{figure*}[h]
\centering
   \includegraphics[height=3in,width=7in]{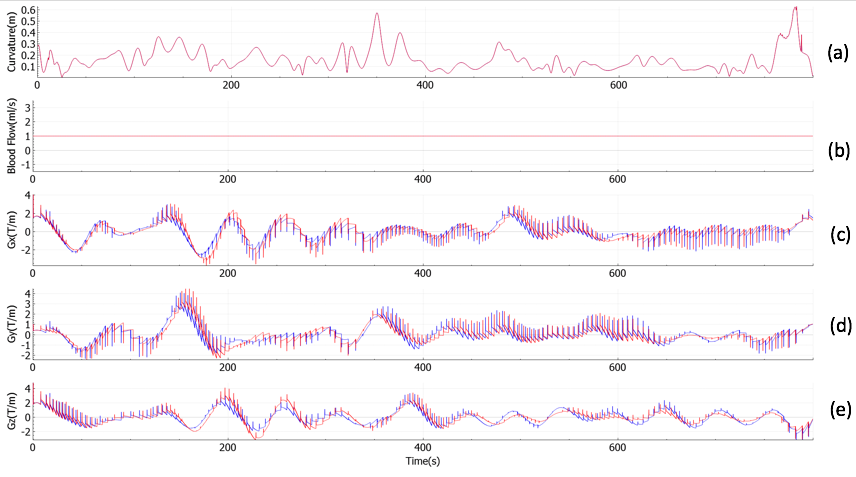}
    \caption{Outcomes for Tp intervals of 100ms (depicted in blue) and 200ms (shown in red) during the simulation of steady blood flow are presented. Figure (a) illustrates the curvature of the path. Figure (b) displays the steady blood flow pattern. Figures (c), (d), and (e) detail the gradient fields produced by the MRI scanner for each respective axis.}\label{GC}
\end{figure*}

\begin{figure*}[h]
\centering
   \includegraphics[height=3in,width=7in]{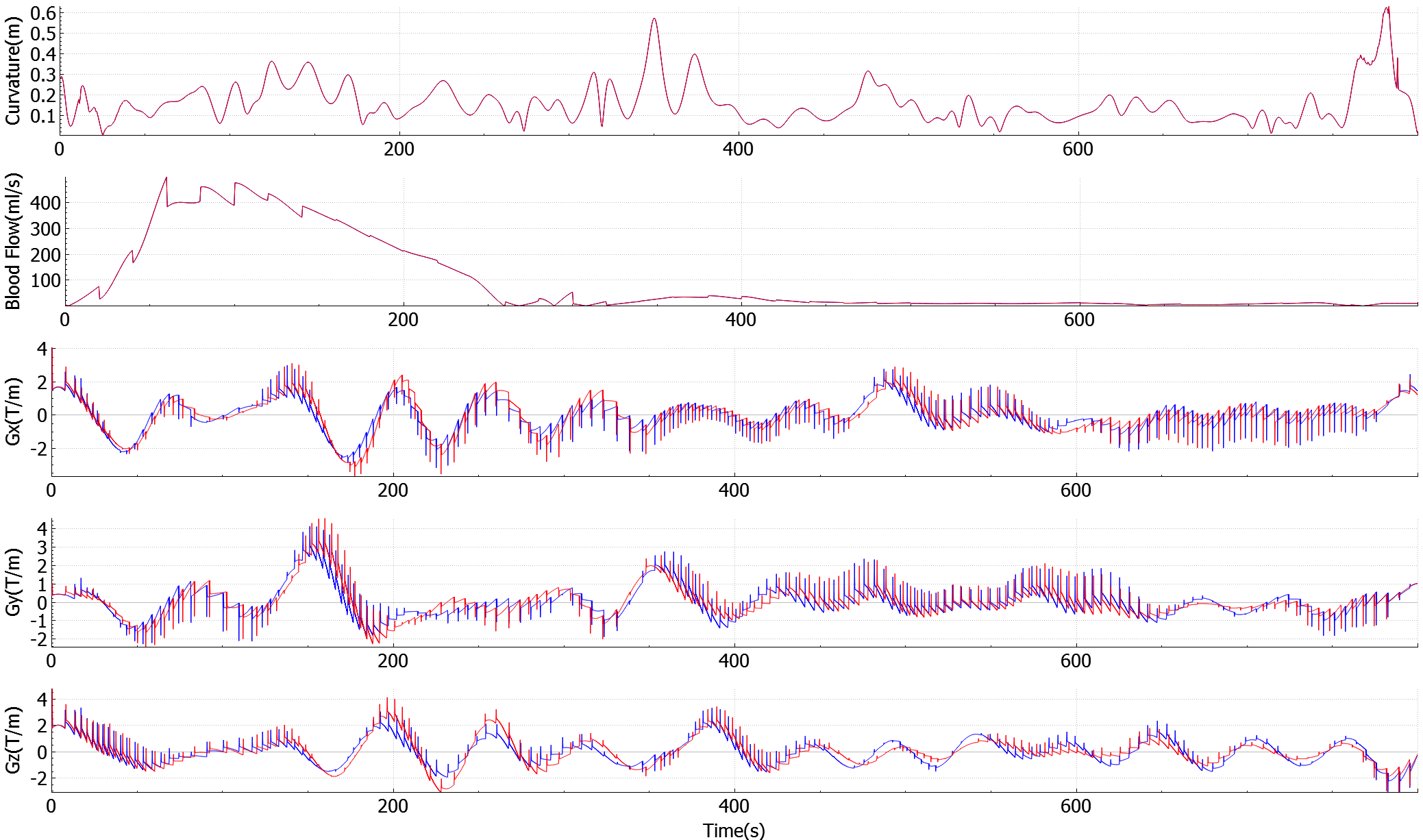}
    \caption{Findings for Tp values of 100ms (in blue) and 200ms (in red) from the simulation under normal blood flow conditions are shown. Figure (a) depicts the curvature of the path. Figure (b) demonstrates the normal blood flow. Figures (c), (d), and (e) display the gradient fields created by the MRI scanner for each axis.}\label{Ga}
\end{figure*}

\begin{figure*}
\centering
   \includegraphics[height=3in,width=7in]{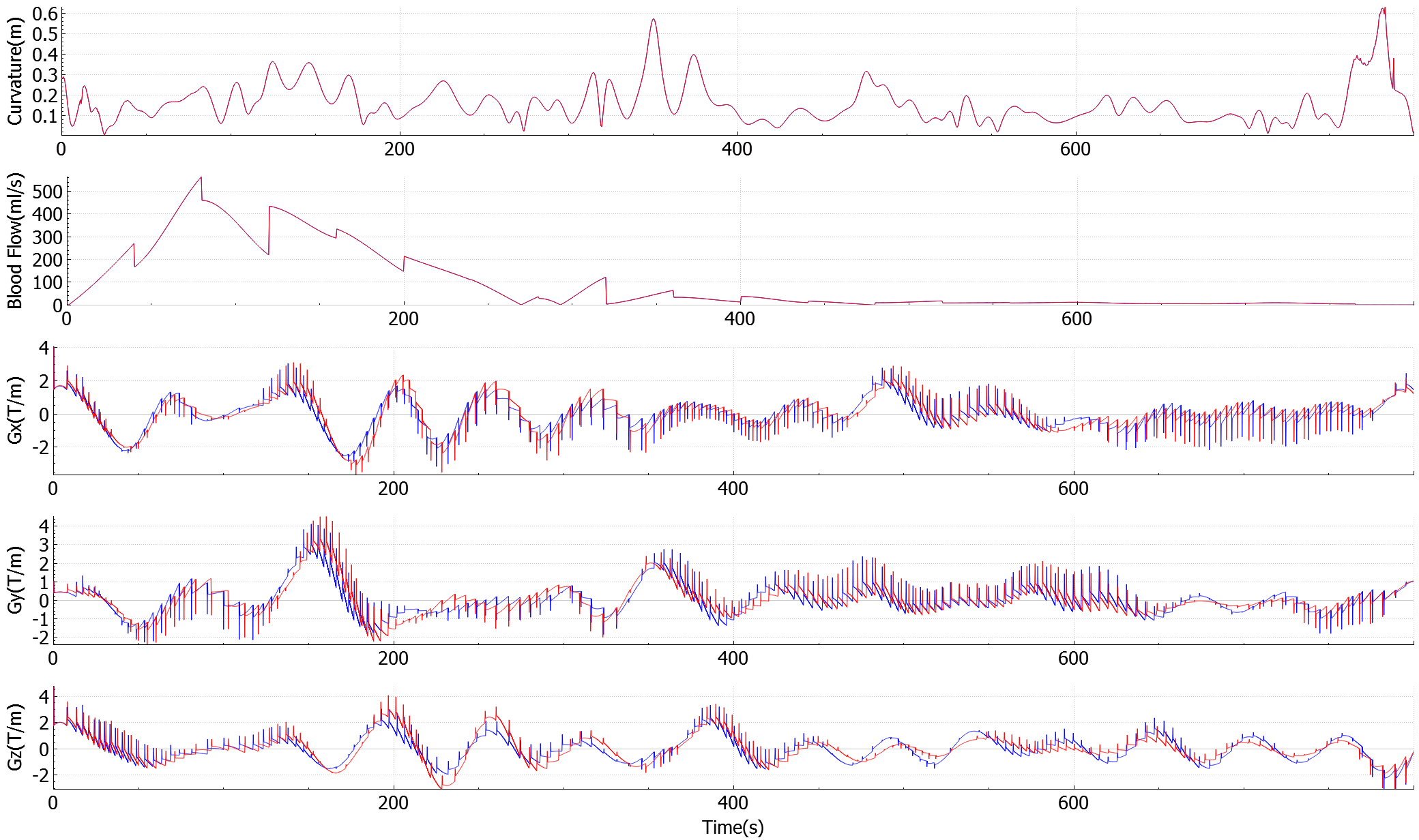}
    \caption{The outcomes from the simulation at Tp intervals of 100ms (illustrated in blue) and 200ms (represented in red) under fast heart rate blood flow conditions are detailed. Figure (a) charts the path's curvature. Figure (b) exhibits the blood flow at a fast heart rate. Figures (c), (d), and (e) outline the gradient fields produced by the MRI scanner for each respective axis.}\label{Gh}
\end{figure*}

Additionally, we include two more variables in our model: $P_c$, representing the MRbot's current location, and $P_s$, the designated position setpoint, where $P_s$ marks a specific point on the MRbot's intended path and $P_c$ reflects its real-time position. Our control system aims to reduce the disparity between $P_c$ and $P_s$, ensuring that the MRbot adheres to its planned trajectory as closely as possible.

By managing these parameters – $V_c$, $V_s$, $K_r$, $P_c$, and $P_s$ – our control system adeptly steers the MRbot through the vascular network. It adjusts its speed and position in real-time, adeptly navigating the human body's complex and dynamic environment. This level of precise control is crucial for the effective implementation of MRbots in medical procedures.

The PID regulator :
\begin{equation}
 \label{eq:pf}
PF= -k_p* Error_v
\end{equation}

\begin{equation}
 \label{eq:pi}
PI= PI-(Error_v* \delta *k_i)
\end{equation}

\begin{equation}
 \label{eq:pd}
PD= -k_d* Error_{dt}
\end{equation} 

\begin{equation}
 \label{eq:error_dt}
Error_{dt} = (Error_v - Error_p) / \delta
\end{equation}

In setting up our PID controller for MRbot navigation, we've chosen specific settings for optimal control, essential for the MRbot's accurate and efficient functioning in the vascular system.

The proportional gain, represented by $k_p$, is established at 2. This figure dictates the PID controller's reaction intensity to current errors, where a higher number indicates a more forceful response. The integral gain, noted as $k_i$, is fixed at 1. It addresses the lingering steady-state error that a solely proportional approach would leave. It accumulates error over time, offering a comprehensive response to counter continuous discrepancies. The derivative gain, $k_d$, is adjusted to 0.01. It anticipates future error trends based on the current error change rate. This foresight helps in averting overshooting and enhances system stability.

The variable $\delta$ signifies the baseline velocity, setting a fundamental speed for the MRbot. Modifying this base velocity enables precise speed adjustments of the MRbot, tailored to specific medical tasks or navigation routes. $Error_p$ tracks the MRbot's past positional errors. Accounting for these historical errors, the PID controller modifies its output to correct deviations from the intended path, ensuring consistent MRbot alignment.

Collectively, these PID controller parameters ($k_p$, $k_i$, $k_d$, $\delta$, and $Error_p$) synergize to create a balanced, responsive control mechanism. This framework adeptly steers the MRbot through the vascular system, adapting to environmental changes and upholding the required trajectory and speed for effective and secure navigation.

\begin{equation}
 \label{eq:moment}
Moment_s = M*Vol
\end{equation} 

\begin{equation}
 \label{eq:vol}
Vol = \frac{4}{3}\times \pi \times r^3
\end{equation}

In our design, various essential parameters define the MRbot's spherical component, crucial for its functionality and guidance. $Moment_s$ signifies the sphere's magnetic moment, a key aspect determining its interaction with magnetic fields. The magnetization of the sphere, noted as ``\textit{M}" with a value of $(1.9496 \times 10^6)$ A/m, is pivotal in gauging the intensity of the magnetic field per unit volume inside the sphere, significantly influencing its reaction to external magnetic forces.

``\textit{Vol}" indicates the sphere's volume. This metric, in conjunction with magnetization, contributes to the sphere's total magnetic moment, influencing its magnetic field interactions.

The sphere's radius, labeled ``\textit{r}," is specified as 0.3mm. This measurement is important as it affects the sphere's volume and surface area, thereby impacting its magnetic characteristics and navigational abilities within the vascular system.

Furthermore, ``\textit{FF}" represents the optimal control linked to gradient adjustments for counteracting the drag force from blood flow on the sphere. This optimal control is essential to ensure that the MRbot efficiently counters the forces it faces in the bloodstream, keeping to its planned trajectory and speed. These parameters together establish the MRbot's sphere's physical and magnetic properties, indispensable for its successful navigation and operation in medical contexts.

\begin{equation}
 \label{eq:ff}
FF=  \frac{1}{2} \cdot C_d \cdot p\cdot Re \cdot V
\end{equation}

In our study, we evaluate critical elements that affect the drag force acting on the MRbot during its journey through blood vessels. The drag coefficient, denoted as $C_{d}$ and valued at 0.47, is a dimensionless measure that quantifies the resistance or drag an object encounters in a fluid. This coefficient is vital for drag force calculation, reflecting the MRbot's shape and flow dynamics in the blood.

The symbol ``\textit{p}" represents blood density, a significant factor in drag force determination. The density of the blood directly impacts the resistance the MRbot faces in this fluid medium.

``\textit{Re}" signifies the sphere's frontal area, a crucial metric for drag force computation. It relates to the MRbot's cross-sectional area that is oriented towards the blood's flow direction, with a larger area resulting in increased resistance or drag force.

Lastly, ``\textit{V}" stands for blood velocity. This velocity is a variable element that varies across different parts of the vascular system. The MRbot's experienced drag force is directly proportional to the blood flow velocity it encounters.

Collectively, these factors - drag coefficient (\textit{Cd)}, blood density \textit{(p)}, sphere's frontal area \textit{(Re)}, and blood velocity \textit{(V)} - are essential for precisely assessing the drag force on the MRbot. Understanding and accurately calculating this force is crucial in designing an MRbot that can navigate efficiently and with control within the human circulatory system.

Gradients \textit{(G)} can be calculated with:
\begin{equation}
 \label{eq:gradient}
G= (1/Moment_s)*(PF+PI+PD+FF)
\end{equation}

$Moment_s$ indicates the sphere's moment, while \textit{PF, PI,} and \textit{PD} correspond to the elements of the PID controller, with \textit{FF} signifying the optimal control. Refer to Figures ~\ref{GC}, ~\ref{Ga}, ~\ref{Gh}, which illustrate noticeable fluctuations in the gradients. These spikes can potentially be attributed to the feedback system of the PID controller. The integral part of the controller, \textit{PI}, requires an initial value for its first computation. This necessity is important even in cases where the scenario of 'i' being zero is circumvented, thus playing a role in the occurrence of these spikes.

To generate a robust magnetic force, it's crucial to increase the magnetization of the MRbot. The process of creating and miniaturizing a ferromagnetic propulsion core is relatively straightforward, especially considering the lower viscosity of blood compared to brain tissue, resulting in a force that scales with the volume of the ferromagnetic component. Our model simplifies the variability in blood vessel diameter, which can change due to heartbeats and impact the drag force, by treating these vessels as inflexible cylindrical tubes. This approach focuses on maximizing magnetization while acknowledging the complexities of the vascular environment.

\begin{equation}
 \label{eq:force}
 F=M \cdot G \cdot {V_{o}}
\end{equation}

Magnetic Resonance Propulsion (MRP) employs magnetic gradients to exert a displacement force on an MRbot. Based on equation (13), \textit{F} symbolizes the magnetic force (\textit{N}) induced by magnetic gradients (T $m^{-1}$), \textit{M} represents the material's magnetization (A/m), ${V_{o}}$ refers to the volume of the ferromagnetic core ($m^3$), and \textit{G} indicates the gradient or the magnetic induction's spatial change. This equation demonstrates that reducing the size or volume of the ferromagnetic core in MRbots, or the ferromagnetic MRbots themselves, leads to a diminished force generated by the MRI system. To achieve adequate magnetic force, employing a ferromagnetic material with a high saturation magnetization is vital. This research does not consider the friction between the sphere and tube walls, as the MRI system's gradients are potent enough to levitate MRbots.

\begin{figure*}[ht!]
\centering
   \includegraphics[height=3in,width=7in]{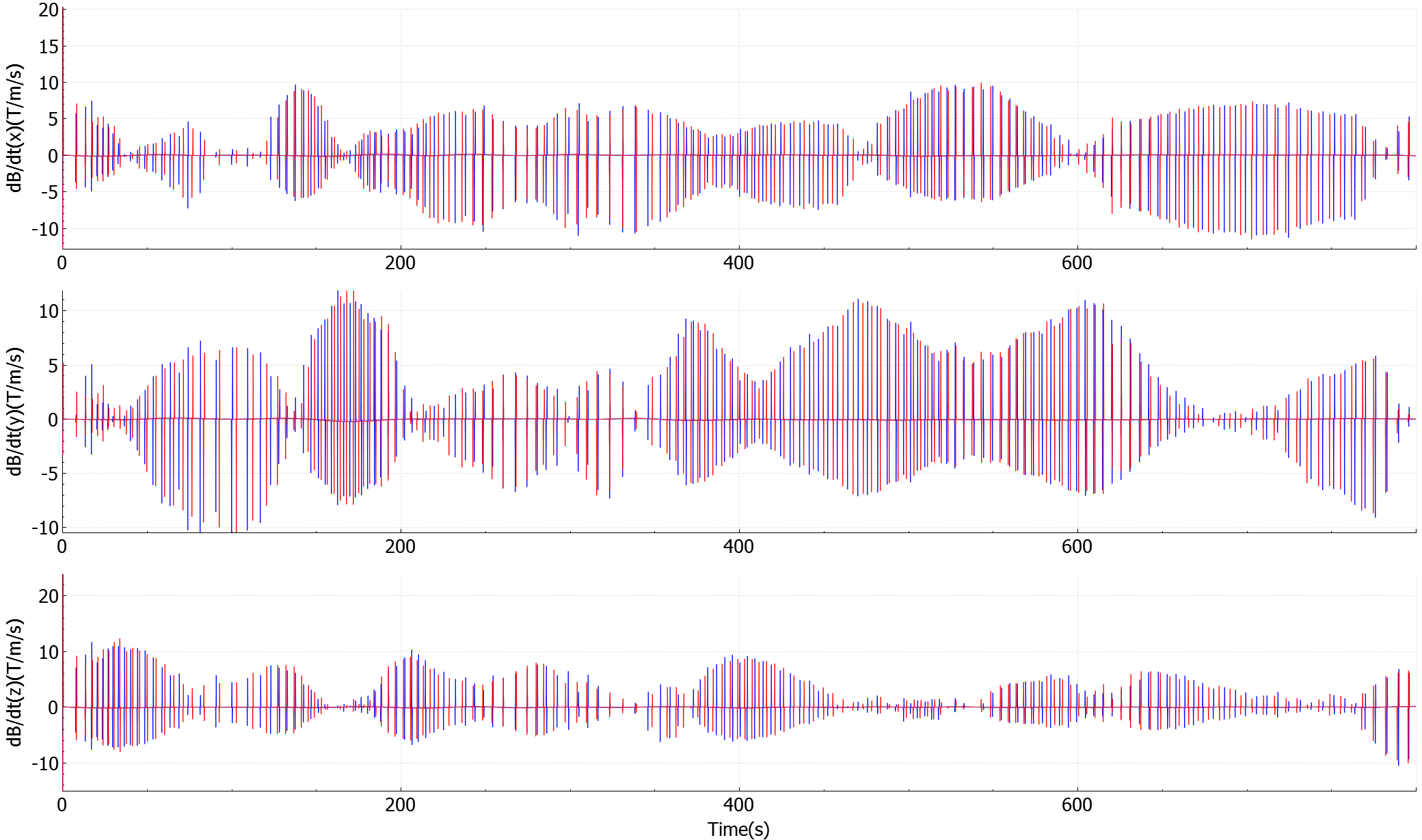}
    \caption{Outcomes for Tp values of 100ms (represented in blue) and 200ms (illustrated in red) were observed in the simulation under steady blood flow conditions. Figures (a), (b), and (c) display the dB/dt rate for each respective axis, with a blood flow rate of 1ml/s.}\label{dC}
\end{figure*}

\begin{figure*}[h]
\centering
   \includegraphics[height=3in,width=7in]{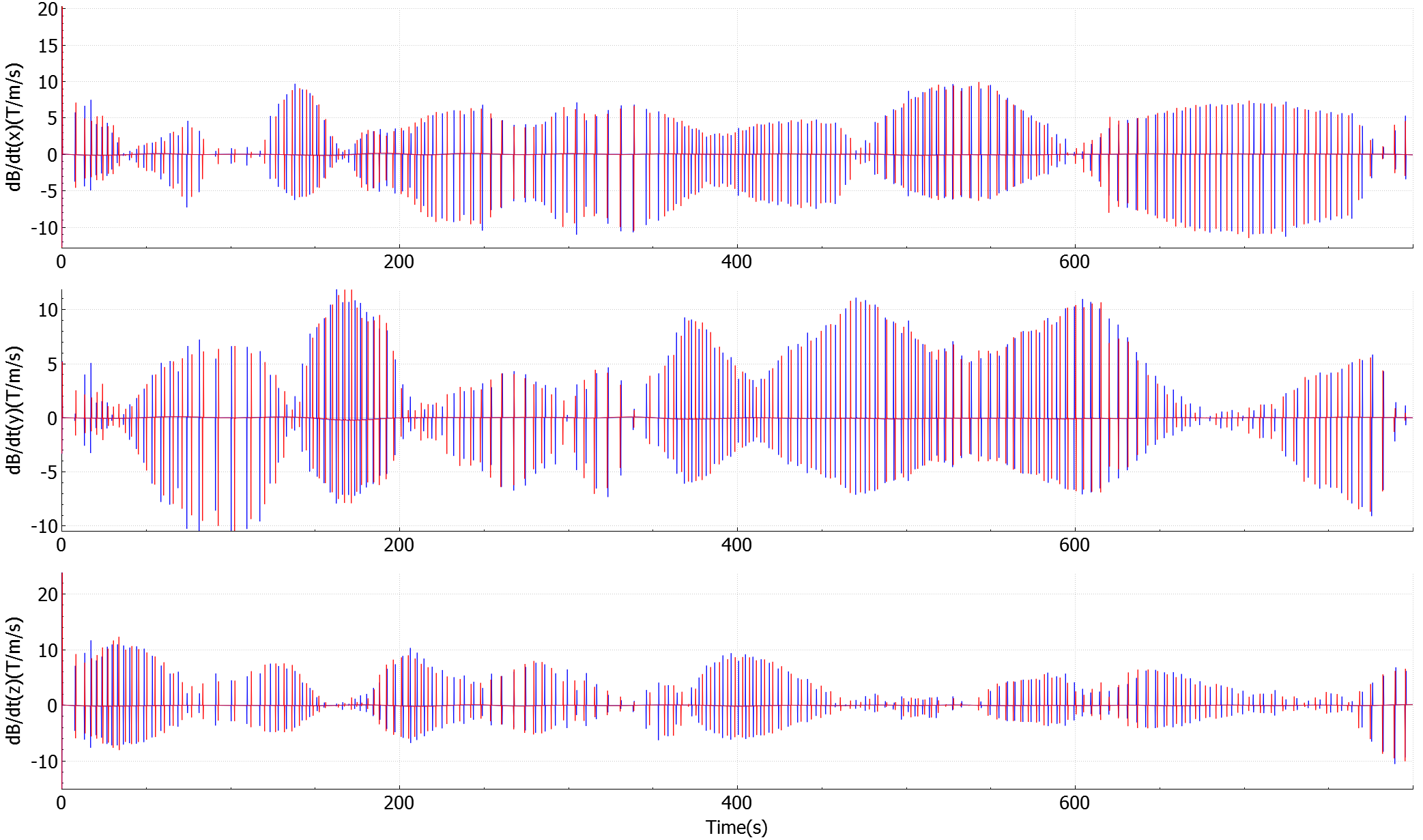}
    \caption{Outcomes from the simulation at Tp settings of 100ms (depicted in blue) and 200ms (shown in red) during normal blood flow conditions were recorded. Charts (a), (b), and (c) depict the dB/dt values for each respective axis in these tests.}\label{Da}
\end{figure*}

\begin{figure*}[h]
\centering
   \includegraphics[height=3in,width=7in]{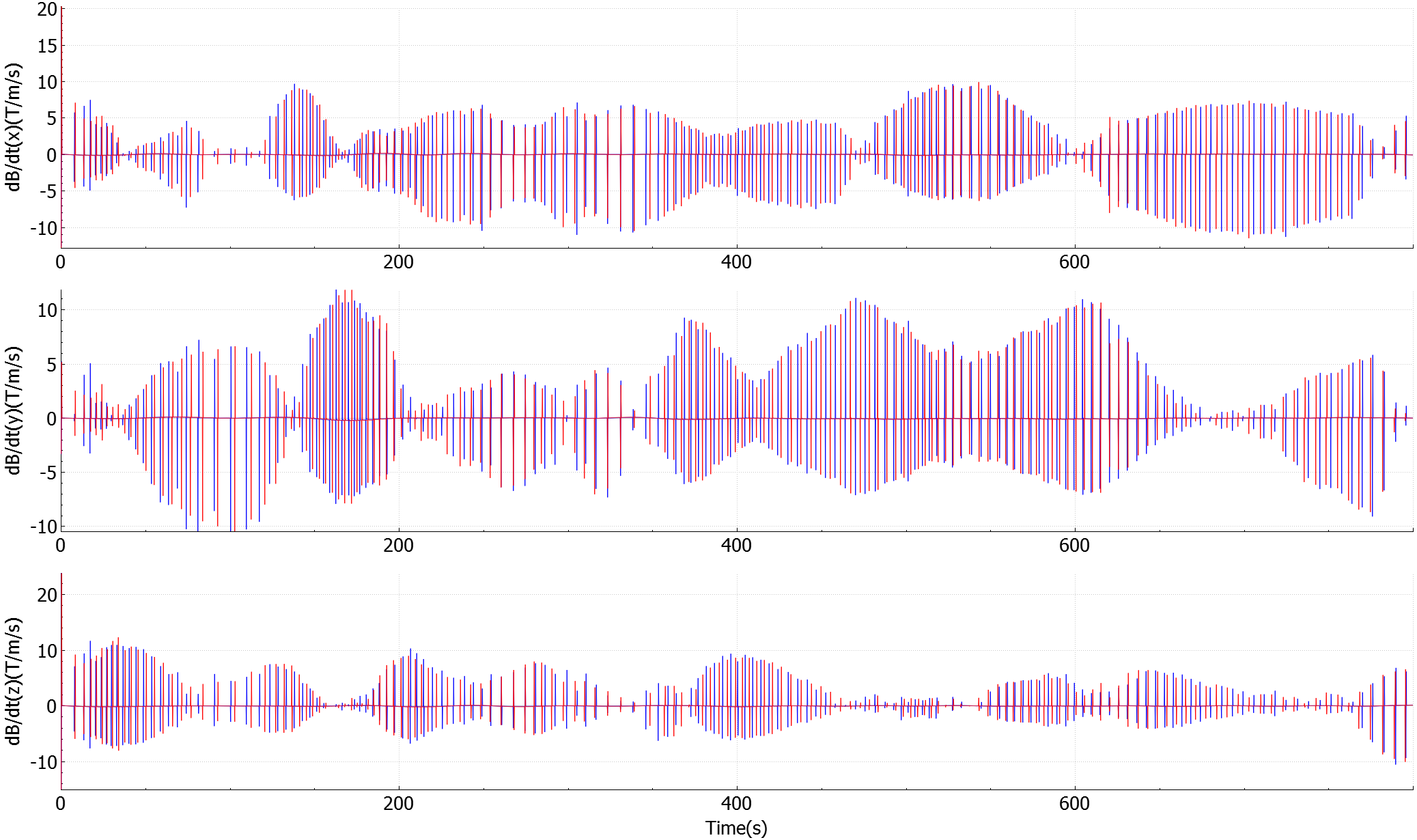}
    \caption{Observations from the simulation with Tp durations of 100ms (blue) and 200ms (red) in the scenario of rapid heart rate blood flow are documented. Graphs (a), (b), and (c) display the dB/dt rates for each axis under these conditions.}\label{Dh}
\end{figure*}

\section{MODELING AND OUTCOMES}

The present magnetic field strengths inbuilt in MRI systems generate gradients up to several tens of mT/m in any direction, creating substantial magnetic forces in varied orientations. To propel the ferromagnetic core along a set path and evaluate this in practice, we examined the influence of MRI system's magnetic field gradients on a ferromagnetic sphere. Materials like permendur, with high magnetization and saturation properties, are optimal for such applications. In our digital model, every position on the 3D vascular path is specified by three gradient coordinates [X(i), Y(i), Z(i)]. The movement of the ferromagnetic sphere hinges largely on the drag force \textit{D} and the propulsion magnetic force \textit{F}. An advanced software framework is key for precise computation and application of gradient data. Our experiment was repeated 1,000 times, with times ranging from 6.8ms to 7.6ms and an average duration of 7.2ms. Additionally, spatial gradients were calculated for more thorough analysis, with values like 0.93 mT/m, 1.68 mT/m, and 3.35 mT/m on the X-axis, and similar varied readings on the Y and Z axes, enabling creation of a magnetic force vector in both upward and downward directions. In our model, we assume constant blood density, meaning the volume of blood entering the vessel is equal to that leaving.

\subsection{Dynamic Magnetic Field Variations}

Gradient coils, typically used in imaging, create time-varying magnetic fields with a certain slew rate, measured in dB/dt. As the magnetic field increases, it may induce an electric current in conductors, including the human body, possibly causing peripheral nerve stimulation, which could be felt as a mild tingle or a brief muscle spasm. However, this is not seen as a major health concern. To avoid potential risks, the FDA has set guidelines for switching gradient fields, especially if the dB/dt is sufficient to propel objects within blood vessels. 

\begin{equation}
\label{eq:slew}
S = G_p/T_r * r
\end{equation}

In this research, \textit{``S"} symbolizes the slew rates (dB/dt), $G_p$ is the gradient difference between two points, $T_r$ is the rise time, and \textit{r} is a 50 cm distance from the isocenter during scanning, with dB/dt measured in Tesla per meter per second (T/m/s). The slew rates and gradient strengths, including x-gradient, y-gradient, and z-gradient, are confirmed to be within FDA's safety limits. For detailed illustrations, refer to Figures \ref{dC}, \ref{Da}, ~\ref{Dh}.

\subsection{Data Simulation to Image Archive Visualization Framework}

Our approach includes enhancing an existing visualization system. The Visualization Toolkit (VTK) is a freely available C++ class library for 3D graphics and visualization. After obtaining the coordinates, we aim to develop a focused object library that integrates seamlessly into our sphere pathway, built of small, distinct segments. Each component in these toolkits is designed to be clearly defined and easy to interface with.

Handling large datasets is an essential requirement. Previous research indicates that ParaView, an open-source, multi-platform visualization application, scales effectively as per [16]. This tool offers a graphical user interface that facilitates the interactive exploration of large datasets. Our system is developed upon ParaView, a C++ based analysis-optimized and modern visualization tool, particularly suited for post-processing in advanced modeling and simulation codes.

VTK forms the cornerstone of ParaView's architecture, providing data representations and algorithms, as well as a mechanism to interconnect these components into a functional program. In our program, red and white lines depict the actual sphere maps and intended paths. The red path includes 8380 experimental data points, charting the sphere’s journey along a selected trajectory. We currently use a step size of 0.0001, which can be adjusted to a value that maintains accuracy without significant loss. This precision allows the MRbot to closely follow the vascular path.

\begin{figure*}
\centering
\includegraphics[width = 16cm,height=8cm]{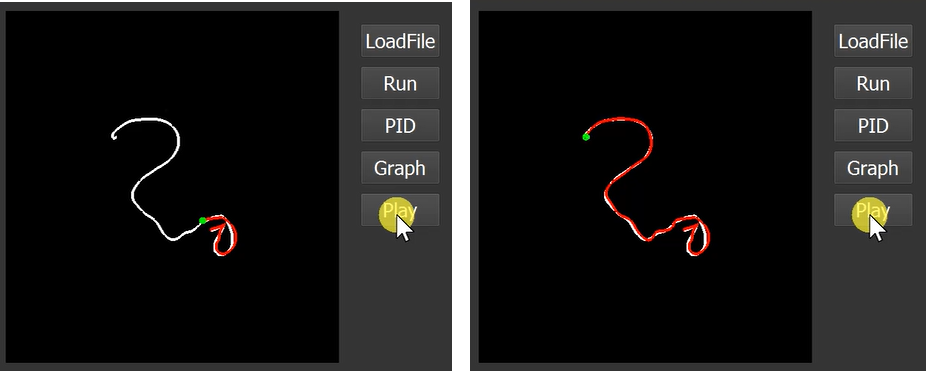}
          \caption{Structure of Qt user interface}
\label{qtt1}
\end{figure*}

Qt stands out as a crucial cross-platform application framework extensively utilized for creating GUI (Graphical User Interface) programs. We employed Qt in the development of web-enabled applications, enabling their deployment across various platforms such as personal computer desktops, mobile devices, and embedded operating systems. This approach allows for the distribution of these applications without the need to rewrite the source code for each platform. See Figure ~\ref{qtt1}. 

\section{DISCUSSION AND CONCLUDING REMARKS}

In our research, we developed a simulation that replicates the movement of a ferromagnetic core through a blood vessel, employing real-time features to accurately mirror real-life physiological scenarios. This research demonstrates the feasibility of designing a Magnetic Resonance Robot (MRbot) that is capable of generating propulsion forces appropriate for navigating the human cardiovascular system. A significant achievement of this study is the successful integration of a tracking method with a sophisticated propulsion system and a Proportional Integral Derivative (PID) controller. This integration allows for the real-time, automatic navigational control of the MRbot within the intricate network of blood vessels.

The development of the core functionalities of this system is carried out in C/C++, ensuring an efficient process from the initial development phase to testing and processing the output. We have incorporated key software libraries which is instrumental in creating detailed 3D graphics and visualizations of the MRbot's path and interactions within the vascular system. Additionally, the simulation includes two realistic blood flow waveforms to simulate the dynamic and complex nature of blood flow within the cardiovascular system more accurately.

A key aspect of our methodology is the utilization of predefined paths for guiding the MRbot, ensuring high-precision propulsion and maneuvering. This precision is vital for the robot's potential medical applications, where accuracy and dependability are essential.  Currently, we have not incorporated virtual routines accessible via graphical user interfaces (GUIs) on LCD screens or augmented reality devices like HoloLens. Integrating these virtual routines would make the control of the MRbot more interactive and user-friendly, potentially broadening its applications and effectiveness. In response, we are developing a theoretical framework that incorporates batch normalization [17][18] and machine learning techniques to predict the performance of these assisted interventions. Batch normalization will help in stabilizing the learning process and improve the performance of our machine learning models, making the system more robust and adaptable to various medical scenarios. This framework aims to provide a deeper understanding of the MRbot's capabilities and limitations, directing future advancements and ensuring the robot's efficient and safe application in medical procedures. The inclusion of machine learning algorithms will enable the MRbot to learn from data, improve its decision-making processes, and adapt to complex medical environments. This approach is expected to enhance the robot's navigation precision, responsiveness to dynamic physiological conditions, and overall effectiveness in therapeutic and diagnostic tasks. This progressive development emphasizes the innovative and impactful nature of our research in the realms of medical robotics and minimally invasive procedures, marking a significant step towards the advancement of intelligent robotic systems in healthcare. Integrating with the aforementioned advancements in medical robotics, our research is also exploring the potential of a generative adversarial network (cGAN) architecture, specifically the S2M-Net model. By simulating realistic three-party conversations, it can provide an immersive learning environment, helping practitioners to better understand the nuances of operating such advanced medical robotics in a safe and controlled setting [19]. 

 %\section*{Acknowledgment}
 %This work was supported by the National Science Foundation award CNS-1646566. All opinions, findings, conclusions or recommendations expressed in this work are those of the authors and do not necessarily reflect the views of our sponsors. 


\begin{thebibliography}{00}


\bibitem{c1}{Huaijuan Zhou, Carmen C, Mayorga-Martinez, Salvador Pané, Li Zhang, and Martin Pumera,} {``Magnetically Driven Micro and Nanorobots," Chemical Reviews, vol. 121, no. 8, pp. 4999-5041, 2021.}  [Online]. Available: https://www.ncbi.nlm.nih.gov/pmc/articles/PMC8154323/


\bibitem{c2} {Wall J, Dhesi J, Snowden C, Swart M,} {``Perioperative medicine Future,”} Healthc J, vol. 9, no. 2, pp. 138-143, July 2022. [Online]. Available: doi: 10.7861/fhj.2022-0051


\bibitem{c3} {Barros AO, Yang J,} {``A Review of Magnetically Actuated Milli/Micro-Scale Robots Locomotion and Features," Crit Rev Biomed Eng, vol. 47, pp. 379-394, 2019.}  [Online]. Available: doi: 10.1615/CritRevBiomedEng.2019030299


\bibitem{c4} Erin O, Boyvat M, Tiryaki M. E, Phelan M,  Sitti M, {``Magnetic Resonance Imaging System-Driven Medical Robotics," Adv. Intell. Syst, vol. 2, pp. 1900110, 2020.}  [Online]. Available: 10.1002/aisy.201900110



\bibitem{c5} Martel S, {``Beyond imaging: Macro- and microscale medical robots actuated by clinical MRI scanners,"  Sci. Robot, vol. 2, 2017.} [Online]. Available: 10.1126/scirobotics.aam8119

\bibitem{c6}  {M. Latulippe and S. Martel}, {``A Progressive Multidimensional Particle Swarm Optimizer for magnetic core placement in Dipole Field Navigation,"} {IEEE/RSJ International Conference on Intelligent Robots and Systems (IROS),} {pp. 2314-2320,} {Oct 2016.} [Online]. Available: https://ieeexplore.ieee.org/document/7759361




\bibitem{c7} Doutel E, Galindo-Rosales FJ, Campo-Deaño L, ``Hemodynamics Challenges for the Navigation of Medical Microbots for the Treatment of CVDs," Materials (Basel), vol. 14, pp. 7402, 2021. [Online]. Available:  doi: 10.3390/ma14237402

\bibitem{c8} {Tucker WD, Arora Y, Mahajan K,}, {``Anatomy, Blood Vessels,"} {StatPearls,} {2023.} [Online]. Available: https://www.ncbi.nlm.nih.gov/books/NBK470401/



\bibitem{c9} {J. B. Mathieu and S. Martel and L. Yahia and G. Soulez and G. Beaudoin,} {``MRI systems as a mean of propulsion for a microdevice in blood vessels,"} {vol. 4,} {pp. 3419-3422,} {Sep 2003.} [Online]. Available: https://ieeexplore.ieee.org/stamp/stamp.jsp?arnumber=1280880

\bibitem{c10} Dyverfeldt, P., Bissell, M., Barker, A.J \textit{et al.} ``4D flow cardiovascular magnetic resonance consensus statement," J Cardiovasc Magn Reson, vol. 17, pp. 72, Aug 2015. [Online]. Available: https://doi.org/10.1186/s12968-015-0174-5 





\bibitem{c11} Shi Y, Lawford P, Hose R, ``Review of zero-D and 1-D models of blood flow in the cardiovascular system," Biomed Eng Online, vol. 11, pp. 33, Apr 26 2011. [Online]. Available:  doi: 10.1186/1475-925X-10-33

\bibitem{c12} Wenhui Chu, Khang Tran, and Nikolaos V. Tsekos, ``Simulations of MRI Guided and Powered Ferric Applicators for Tetherless Delivery of Therapeutic Interventions," In Proceedings of the 2022 12th International Conference on Bioscience, Biochemistry and Bioinformatics (ICBBB '22), pp. 29-37, 2022.
 [Online]. Available: https://doi.org/10.1145/3510427.3510432

\bibitem{c13} {Papanastasiou, T., Georgiou, G.,  Alexandrou, A.N,}{ ``Viscous Fluid Flow (1st ed.),"} {CRC Press, 1999}. [Online]. Available: https://doi.org/10.1201/9780367802424


\bibitem{c14} R. Kehlenbeck and R. Di Felice, {``Empirical relationships for the terminal settling velocity of sheres in cylindrical columns,"} Chem. Eng. Technol., vol. 21, pp. 303, 1999. [Online]. Available: https://www.semanticscholar.org/paper/

\bibitem{c15} {M. Sitti, H. Ceylan, W. Hu, J. Giltinan, M. Turan, S. Yim, E. Diller,} {``Biomedical Applications of Untethered Mobile Milli/Microrobots,"} {Vol. 103,} {pp. 205-224,} {2015.} [Online]. Available: https://pubmed.ncbi.nlm.nih.gov/27746484/

\bibitem{c16} Ahrens, J Geveci, Berk Law, Charles, ``ParaView: An End-User Tool for Large Data Visualization," Visualization Handbook, 2005. [Online]. Available: https://www.semanticscholar.org/paper/

\bibitem{c17}W. Chu \textit{et al.} ``BNU-Net: A Novel Deep Learning Approach for LV MRI Analysis in Short-Axis MRI," 2019 IEEE 19th International Conference on Bioinformatics and Bioengineering (BIBE), pp. 731-736, Athens, Greece, 2019.  [Online]. Available: https://ieeexplore.ieee.org/document/8941959

\bibitem{c18}Wenhui Chu and Nikolaos V. Tsekos, ``Two Deep Learning Approaches for Automated Segmentation of Left Ventricle in Cine Cardiac MRI," In Proceedings of the 2022 12th International Conference on Bioscience, Biochemistry and Bioinformatics, pp. 7-13, 2022. [Online]. Available: https://doi.org/10.1145/3510427.3510429

\bibitem{c19} Aobo Jin, Qixin Deng, and Zhigang Deng,  ``S2M-Net: Speech Driven Three-party Conversational Motion Synthesis Networks," In Proceedings of the 15th ACM SIGGRAPH Conference on Motion, Interaction and Games (MIG '22), Association for Computing Machinery, New York, NY, USA, Article 2, 2022. [Online]. Available: https://doi.org/10.1145/3561975.3562954 




%\bibitem{c25}

%\bibitem{c26}

%\bibitem{c27}

%\bibitem{c28}

%\bibitem{c5}J. Cong, Y. Zheng, W. Xue, B. Cao and S. Li, ``MA-Shape: Modality Adaptation Shape Regression for Left Ventricle Segmentation on Mixed MR and CT Images," in IEEE Access, vol. 7, pp. 16584-16593, 2019. doi: 10.1109/ACCESS.2019.2892965. [Online]. Available: https://ieeexplore.ieee.org/abstract/document/8611328

%\bibitem{c6}Margeta, J., Geremia E., Criminisi A., Ayache N,  ``Layered spatio temporal forests for left ventricle segmentation from 4D cardiac MRI data." MICCAI workshop: Statistical Atlases and Computational Models of the Heart (STACOM). 2012.  [Online]. Available: https://www.microsoft.com/en-us/research/publication/layered-spatio-temporal-forests-for-left-ventricle-segmentation-from-4d-cardiac-mri-data/
%\bibitem{c90}Shengfeng Liu, Yi Wang, Xin Yang, Baiying Lei, Li Liu, Shawn Xiang Li, Dong Ni, Tianfu Wang, "Deep Learning in Medical Ultrasound Analysis: A Review,Engineering," Vol 5, Issue 2, 2019, pp 261-275, ISSN 2095-8099,
%https://doi.org/10.1016/j.eng.2018.11.020. [Online]. Available: http://www.sciencedirect.com/science/article/pii/S2095809918301887
%\bibitem{c7}X.Zhou and G.Yang,  ``Normalization  in  Training  U-Net for 2-D Biomedical Semantic Segmentation," in IEEE Robotics and Au- tomation Letters, vol. 4, no. 2, pp. 1792-1799, April 2019. doi: 10.1109/LRA.2019.2896518. [Online]. Available: https://arxiv.org/abs/1809.03783

%\bibitem{c8}Ronneberger, Olaf et al. ``U-Net: Convolutional Networks for Biomedical Image Segmentation.” ArXiv abs/1505.04597 (2015). [Online]. Available: https://arxiv.org/abs/1505.04597

%\bibitem{c88}Ioffe, Sergey and Szegedy, Christian. ``Batch Normalization: Accelerating Deep Network Training by Reducing Internal Covariate Shift.." Paper presented at the meeting of the ICML, 2015. [Online]. Available: https://dl.acm.org/citation.cfm?id=3045118.3045167

%\bibitem{c9} Phi Vu Tran, ``A Fully Convolutional Neural Network for Cardiac Segmentation in Short-Axis MRI,"  2016.
%[Online]. Available: https://arxiv.org/abs/1604.00494

%\bibitem{c89}Croitoru, Ioana and Bogolin, Simion-Vlad
%and Leordeanu, Marius. ``Unsupervised Learning of Foreground Object Segmentation." vol 127,
%no 9, pp.1279-1302  https://doi.org/10.1007/s11263-019-01183-3
%[Online]. Available: https://link.springer.com/article/

%\bibitem{c10} P.Y.Simard, D.Steinkraus and J. C. Platt, ``Best practices for convolutional neural networks applied to visual document analysis," Seventh International Conference on Document Analysis and Recognition, Proceedings., Edinburgh, UK, 2003, pp. 958-963.
%doi: 10.1109/ICDAR.2003.1227801. [Online]. Available: https://ieeexplore.ieee.org/document/1227801

%\bibitem{c11} Ngo, T.A., G. Carneiro. ``Left ventricle segmentation from cardiac MRI combining level set methods with deep belief networks." 20th IEEE International Conference on Image Processing. pp. 695-699, 2013. [Online]. Available: https://ieeexplore.ieee.org/document/6738143

%\bibitem{c12} Hu, H., Liu H., Gao Z., Huang L.  ``Hybrid segmentation of left ventricle in cardiac MRI using gaussian-mixture model and region restricted dynamic programming." vol. 31, 5 2013, pp. 575-584. [Online]. Available: https://www.sciencedirect.com/science/article/pii/S0730725X12003670

%\bibitem{c13} Liu, H., Hu H., Xu X., Song E., 2012. ``Automatic Left Ventricle Segmentation in Cardiac MRI Using Topological Stable-State Thresholding and Region Restricted Dynamic Programming." Academic Radiology, vol 19, no. 6, pp. 723 - 731. 2012. [Online]. Available:
% https://doi.org/10.1016/j.acra.2012.02.011
 
%\bibitem{c18} X. Gao, N. V. Navkar, D. J. Shah, N. V. Tsekos and Z. Deng, ``Intraoperative registration of preoperative 4D cardiac anatomy with real-time MR images," 2012 IEEE 12th International Conference on Bioinformatics Bioengineering (BIBE), Larnaca, 2012, pp. 583-588.
%doi: 10.1109/BIBE.2012.6399737 [Online]. Available:
%https://ieeexplore.ieee.org/document/6399737



\end{thebibliography}
\end{document}